\documentclass[11pt]{article}

\usepackage[preprint]{acl}

\usepackage{times}
\usepackage{latexsym}
\usepackage{booktabs}
\usepackage{natbib} %
\usepackage{times}
\usepackage{pifont}
\usepackage{graphicx}
\usepackage{booktabs}
\usepackage{xcolor}
\usepackage{amsmath, amsthm, amssymb, amsfonts}
\usepackage{thmtools}
\usepackage{graphicx}
\usepackage{setspace}
\usepackage{float}
\usepackage{hyperref}
\usepackage[utf8]{inputenc}
\usepackage[english]{babel}
\usepackage{framed}
\usepackage{cleveref}
\usepackage[inline]{enumitem}
\usepackage{hyperref}
\usepackage{url}
\usepackage{wrapfig}
\usepackage{subcaption}
\usepackage{comment}
\usepackage{multirow}
\usepackage[textsize=tiny]{todonotes}
\usepackage{listings}
\usepackage{caption}
\usepackage{float}
\usepackage{graphicx}

\graphicspath{/figures/}

\newcommand{\mergekit}{\texttt{MergeKit}}
\newcommand{\pymoo}{\texttt{PyMoo}}
\newcommand{\method}[1]{\texttt{\textsf{#1}}}

\usepackage[T1]{fontenc}

\usepackage[utf8]{inputenc}

\usepackage{microtype}

\usepackage{inconsolata}

\usepackage{graphicx}

\newcommand{\tubeemoji}{\includegraphics[height=1em]{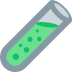}}

\newcommand{\greencheck}{{\color{teal}\ding{51}}}
\newcommand{\redcross}{{\color{purple}\ding{55}}}
\newcommand{\mergenetic}{\texttt{Mergenetic}} %

\title{\tubeemoji{}  \textit{\mergenetic{}}: a Simple Evolutionary Model Merging Library}

\author{Adrian Robert Minut\textsuperscript{1$^\star$}, Tommaso Mencattini\textsuperscript{2$^\star$}, Andrea Santilli\textsuperscript{1}, Donato Crisostomi\textsuperscript{1}, \\ \textbf{Emanuele Rodolà\textsuperscript{1}}\\
  \textsuperscript{1}Sapienza University of Rome 
  \hspace{0.24cm}
  \textsuperscript{2}Ecole Polytechnique Fédérale de Lausanne 
  \hspace{0.2cm}\\
  \texttt{minut@di.uniroma1.it}
}

\newcommand{\lmeval}{\texttt{LM-Eval-Harness}}

\usepackage[most]{tcolorbox}
\newtcbox{\code}{on line,
  colback=gray!15,
  colframe=gray!60,
  boxrule=0pt,
  boxsep=1pt,
  left=3pt,
  right=3pt,
  top=1pt,
  bottom=1pt,
  arc=2pt,
  fontupper=\ttfamily\footnotesize,
  valign=center,
}

\newcommand\blfootnote[1]{%
  \begingroup
  \renewcommand\thefootnote{}\footnote{#1}%
  \addtocounter{footnote}{-1}%
  \endgroup
}

\usepackage{emoji}

\begin{document}

\maketitle

\begin{abstract}
Model merging allows combining the capabilities of existing models into a new one---post hoc, without additional training.\blfootnote{$\star$ denotes equal contribution.}
This has made it increasingly popular thanks to its low cost and the availability of libraries that support merging on consumer GPUs. Recent work shows that pairing merging with evolutionary algorithms can boost performance, but no framework currently supports flexible experimentation with such strategies in language models.
    We introduce Mergenetic\footnote{\href{https://github.com/tommasomncttn/mergenetic}{https://github.com/tommasomncttn/mergenetic}}, an open-source library for evolutionary model merging. %
    Mergenetic enables easy composition of merging methods and evolutionary algorithms, while incorporating lightweight fitness estimators to reduce evaluation costs. We describe its design and demonstrate that Mergenetic produces competitive results across tasks and languages using modest hardware. 
    %A video demo showcasing its main features is also provided\footnote{\href{https://youtu.be/lazoVeP7ku8}{https://youtu.be/lazoVeP7ku8}}.
\end{abstract}

\section{Introduction}
Recent advances in large language models (LLMs) have shown that merging previously fine-tuned models can yield new systems with complementary strengths --- often surpassing any single constituent \cite{yang2024model}. Rather than fully re-training from scratch or fine-tuning a large foundation model for every new task, merging techniques compose knowledge that is already encoded in existing checkpoints (e.g., specialized domain knowledge, multilingual abilities, or skills).

The accessibility of model merging has expanded significantly due to its inexpensive nature coupled with easy-to-use libraries like \mergekit{} \cite{mergekit}, enabling practitioners to produce competitive models from existing ones using standard consumer GPUs. Indeed, at the time of writing, approximately 30\% of models on the Hugging Face Open LLM leaderboard \cite{open-llm-leaderboard-v2} are merged models \citep{task-vectors}.

Recent research has shown that combining model merging with evolutionary algorithms can achieve superior performance \cite{sakana,merge3}. However, this approach faces two key challenges: first, there is currently no library for experimenting with different evolutionary algorithms and merging methods; second, these methods typically require repeated evaluations on an evolutionary datasets to compute fitness functions, making them more computationally expensive than standard merging techniques. These limitations restrict access for the very user base that model merging was intended to empower.

\begin{figure}[t]
    \centering
    \includegraphics[width=\linewidth]{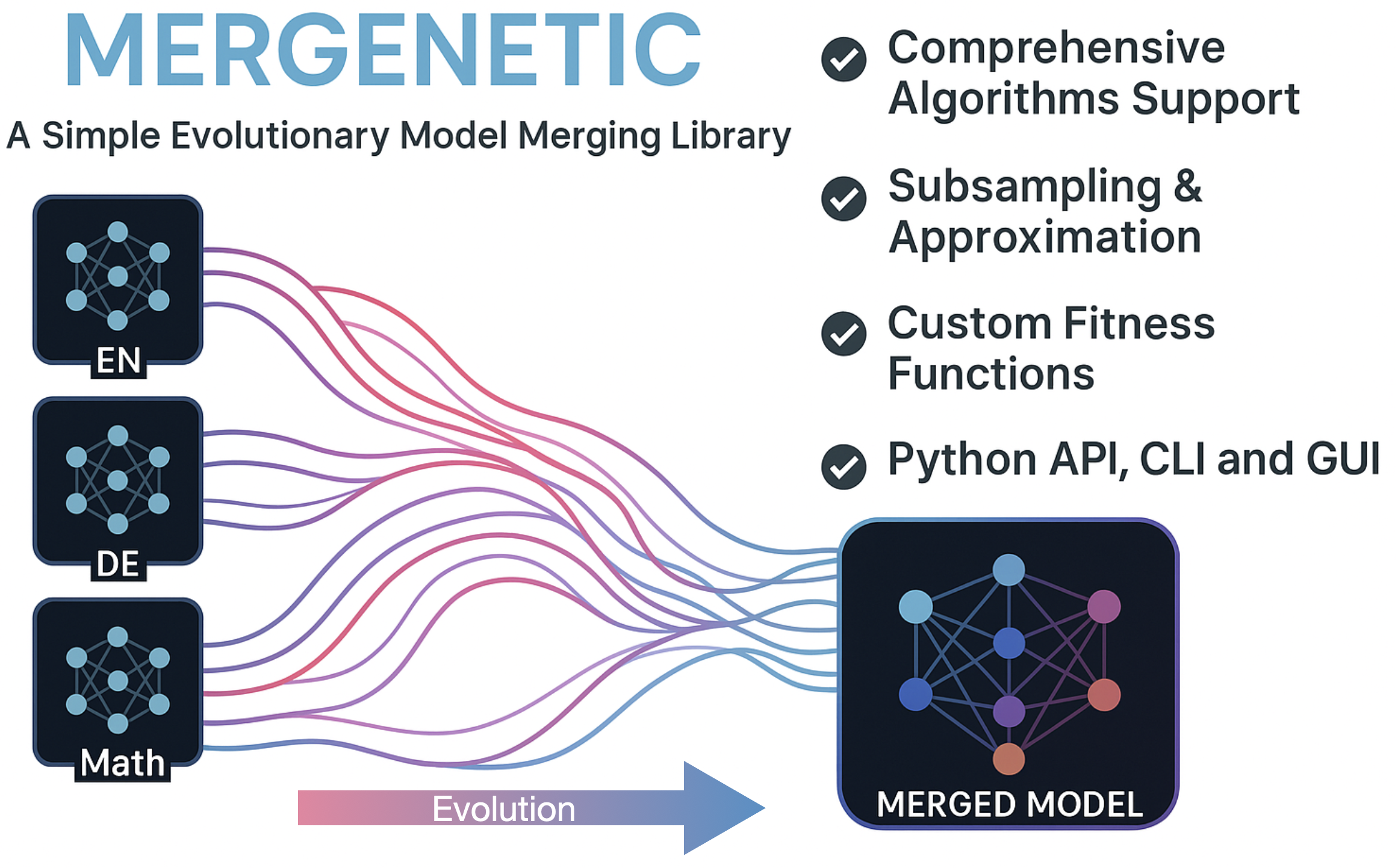}
    \caption{\mergenetic{} makes it easy to produce new state-of-the-art LLMs with minimal requirements.}
    \label{fig:teaser}
\end{figure}

In this paper, we introduce \emph{\mergenetic{}}, a simple library to easily perform evolutionary model merging.
Built on top of \mergekit{} \cite{mergekit} and the widely used evolutionary framework \pymoo{} \cite{pymoo}, our library provides:
\begin{enumerate}
     \item \textbf{Comprehensive Algorithm Support.}
    \mergenetic{} integrates 19 evolutionary algorithms and 6 merging strategies, enabling both single- and multi-objective optimization. This includes classical methods like genetic algorithms and state-of-the-art approaches such as \method{NSGA-II} \cite{nsga2}.
    
    \item \textbf{Subsampling \& Approximation.} 
    To reduce the overhead of fitness evaluations and support merging on consumer GPUs, \mergenetic{} allows for selective evaluation over dataset subsets and supports advanced approximation techniques for efficient fitness estimation \cite{merge3, tinybenchmarks}.
    
    \item \textbf{Custom Fitness Functions.} The library seamlessly integrates with \texttt{LM-Eval-Harness}\footnote{\href{https://github.com/EleutherAI/lm-evaluation-harness}{github.com/EleutherAI/lm-evaluation-harness}} \cite{lm-eval}, offering out-of-the-box support for 8000+ tasks and metrics for fitness computation. Users can also define their own fitness routines tailored to specific needs.

    \item \textbf{Python API, CLI, and GUI.} 
    \mergenetic{} provides a flexible Python API for power users who wish to customize merging workflows, alongside a command-line interface (CLI) and a graphical user interface (GUI) for quick and intuitive setup. Through the CLI or GUI, users can select models from the Hugging Face Hub, configure fitness functions, and launch merging experiments without writing code.
\end{enumerate}

Figure~\ref{fig:teaser} and Table~\ref{tab:comparison_mergekit} summarize the key features of the library.
By making evolutionary model merging more \emph{efficient}, \emph{configurable}, and \emph{accessible}, \mergenetic{} expands the potential of merging as a truly \emph{democratizing} technique.

In the remainder of this paper, we describe (i) the relevant background for \mergenetic{}, (ii) comparisons with existing solutions, (iii) its system architecture and workflow, and (iv) empirical evaluations featuring cross-lingual math merges and multi-task merges on publicly available LLMs. Finally, we conclude by discussing future extensions and potential broader impacts of this approach.

\section{Background and Related Work}

\definecolor{codegreen}{rgb}{0,0.6,0}
\definecolor{codegray}{rgb}{0.5,0.5,0.5}

\definecolor{backcolour}{RGB}{245,248,250}
\definecolor{emph}{RGB}{166,88,53}
\definecolor{nightblue}{RGB}{9,49,105}
\definecolor{keywords}{RGB}{207,33,46}
\definecolor{lightpurple}{RGB}{130,81,223}

\lstdefinestyle{mystyle}{
    backgroundcolor=\color{backcolour},   
    commentstyle=\color{codegreen},
    keywordstyle=\color{keywords},
    stringstyle=\color{nightblue},
    basicstyle=\ttfamily\footnotesize,
    breakatwhitespace=false,         
    breaklines=true,                 
    captionpos=b,                    
    keepspaces=true,                 
    showspaces=false,                
    showstringspaces=false,
    showtabs=false,                  
    tabsize=2,
    frame=shadowbox,
    emph={LinearMerger, MergingProblem, GA, Searcher},
    emphstyle={\color{emph}},
    emph={[2]search,test},
    emphstyle={[2]\color{lightpurple}}
}

\lstset{style=mystyle}

\begin{figure*}[!t]
    \centering
    \begin{lstlisting}[language=Python]
from mergenetic.merging.linear_merger import LinearMerger
from mergenetic.optimization.merging_problem import MergingProblem
from pymoo.algorithms.soo.nonconvex.ga import GA
from mergenetic.searcher import Searcher

# Initialize the merger with base model, finetuned models, and output paths
merger = LinearMerger(run_id="demo_run", 
                      path_to_base_model="my/base/model",
                      model_paths=["finetunedA", "finetunedB"], 
                      path_to_store_yaml="configs/merging_config.yaml", 
                      path_to_store_merged_model="merged_checkpoints/",  
                      dtype="float16")

# Define the optimization problem for merging
problem = MergingProblem(
  merger    = merger,        # Merger object
  search_df = my_dev_data,   # Dataset used to compute fitness
  n_var     = 2,             # Number of variables (weights for the models)
  n_obj     = 1              # Number of objectives (usually a single metric)
)

algorithm = GA(pop_size=10)  # Genetic algorithm with population size 10

# Create searcher to run GA over the merging problem
searcher = Searcher(problem, algorithm, results_path="results/",
                    n_iter=50, seed=42, run_id="demo_run")

searcher.search()  # Run the evolutionary search for optimal weights
searcher.test()    # Evaluate the final merged model
\end{lstlisting}
    \caption{Example on how to use the Python API for power users who wish to customize merging workflows.}
    \label{fig:example_code}
\end{figure*}

\paragraph{Model Merging.}
Model merging \citep{git-rebasin, cycle-consistent, rebasin-implicit-sinkhorn, task-vectors, ties, yu2024language, matenamerging, wortsman2022model, zip-it} has become a powerful and efficient alternative to ensembling, enabling the integration of existing models without requiring additional training. \mergenetic{} focuses on the multi-task scenario, where the aim is to merge different fine-tunings of a single pretrained model \citep{task-vectors, ties, yu2024language, matenamerging, wortsman2022model, davari2023model, wang2024localizing, zhou2024atm, gargiulo2025tasksingularvectorsreducing, sakana, choshen2022fusing}. 

\paragraph{Evolutionary Algorithms.}
Evolutionary Algorithms (EAs) are black-box optimization techniques that operate on a population of candidate solutions, evolving them over successive generations using operators such as selection, mutation, recombination, and crossover \citep{6791438, petrowski2017evolutionary, dasgupta1997evolutionary, real2019regularized, vincent2023improved}. 
A key component of EAs is the \textit{fitness function}, which quantifies the quality of each candidate and steers the evolutionary process by promoting higher-performing solutions \citep{ea}. Applying EAs to model merging, evolutionary merging techniques \citep{sakana, merge3} automatically search for effective merging recipes using the performance of the merged model on a held-out validation dataset as the fitness function.

\paragraph{Comparison with other libraries.}
\begin{table}[t]
\centering
\resizebox{\columnwidth}{!}{%
\begin{tabular}{lcc}
\toprule
\textbf{Features} & \textbf{\mergenetic{} (Ours)} & \textbf{MergeKit} \\
\midrule
Merging Algorithms & \textbf{6} & 5\footnotemark \\
Evolutionary Algorithms & \textbf{19} & 1 \\
Multi-objective & \greencheck & \redcross \\
Dataset Subsampling & \greencheck~(Random + Custom) & \redcross \\
Custom Fitness Functions & \greencheck & \redcross \\
GUI & \greencheck & \redcross \\
\bottomrule
\end{tabular}
}
\caption{Comparison of \mergenetic{} and \mergekit{}.}
\label{tab:comparison_mergekit}
\end{table}
\footnotetext{This number refers to the supported merging methods in evolutionary merging as per the \href{https://github.com/arcee-ai/mergekit/blob/main/docs/evolve.md}{documentation}.}
The most closely related library to \mergenetic{} is \mergekit{} \cite{mergekit}, which provides the underlying merging strategies (e.g., \method{TIES}, \method{DARE}, \method{SLERP}) that we build upon in our evolutionary pipelines. However, when it comes to search capabilities, \mergekit{} supports only a single evolutionary algorithm --- CMA-ES \cite{cmaes} --- offering limited flexibility in exploring the optimization landscape. In contrast, \mergenetic{} integrates with the full suite of algorithms from \texttt{pymoo}, enabling users to choose from a broad range of single- and multi-objective evolutionary strategies, as shown in Table~\ref{tab:optimization_algorithms}.

Most importantly, \mergekit{} assumes that the fitness function must be computed over the full evaluation dataset, which significantly increases runtime and computational demands --- often making the entire process impractical on consumer hardware. In contrast, \mergenetic{} supports sub-sampled evaluation and advanced fitness estimation techniques (e.g., IRT-based estimators \cite{tinybenchmarks, merge3}), dramatically reducing evaluation cost and enabling high-quality merging to be performed efficiently on a single GPU.

\begin{table}[t]
\centering
\resizebox{\columnwidth}{!}{%
\begin{tabular}{lcc}
\toprule
\textbf{Supported Merging Method} & \textbf{Multi-Model} & \textbf{Base Model} \\
\midrule
\method{Task Arithmetic} \cite{taskari} & \greencheck & \greencheck \\
\method{Model Soups} \cite{wortsman2022model} & \greencheck & \redcross \\
\method{SLERP}  & \redcross & \greencheck \\
\method{TIES} \cite{ties} & \greencheck & \greencheck \\
\method{DARE} \cite{yu2024language} + \method{TIES} & \greencheck & \greencheck \\
\method{DARE} \cite{yu2024language} + \method{Task Arithmetic} & \greencheck & \greencheck \\
\bottomrule
\end{tabular}}
\caption{Supported merging methods in \mergenetic}
\label{tab:merge-methods}
\end{table}

\section{Design and guiding principles}
\label{sec:design}

The design of \mergenetic{} reflects our goal of supporting a wide range of evolutionary model-merging experiments on consumer hardware. In this section, we outline the guiding principles that drove our design decisions before diving into key modules and functionalities in \S\ref{sec:modules}. 

\paragraph{Research-Oriented}
\label{sssec:research_oriented}

A central motivation for \mergenetic{} is to enable \emph{researchers} to easily explore and compare different evolutionary algorithms, merging strategies, and optimization objectives. Rather than locking users into a fixed routine, \mergenetic{} supports a flexible mix of merging methods (e.g., \method{TIES}, \method{DARE}, \method{SLERP} from \mergekit{} \cite{mergekit}), evolutionary algorithms (e.g., GA, NSGA-II, DE from \pymoo{} \cite{pymoo}), and evaluation backends (e.g., LM-Eval-Harness or user-defined). This modularity supports systematic experimentation, such as comparing single- vs.\ multi-objective merges or testing different data sampling strategies—and allows defining custom objectives through simple subclassing.

\paragraph{User-Friendly} \label{sssec:consumer_friendly} To democratize model merging for researchers and practitioners with standard GPU setups, \mergenetic{} is designed to be both configuration-centric and user-friendly. Users can define merges, tasks, algorithms, and evaluators using simple YAML files, a command-line interface, or an interactive GUI --- minimizing the engineering overhead typical of large-scale experiments. The library is optimized for consumer GPUs by supporting approximate evaluation methods (e.g., IRT-based estimators), dataset sub-sampling, and partial model loading. It integrates seamlessly with \lmeval{}, supporting over 8000+ tasks and metrics already defined in the library (e.g., GSM8K and ARC), while also making it easy to plug in custom datasets and evaluations for fitness computation. Together, these features enable meaningful evolutionary merging on a single GPU, lowering the barrier for smaller labs and individual practitioners.

\section{Mergenetic}
\paragraph{Modules and Functionalities}
\label{sec:modules}

The implementation relies on \mergekit{} \cite{mergekit} for merging the models, \pymoo{} \cite{pymoo} for optimizing the objective function through evolutionary algorithms, and \lmeval{} \cite{lm-eval} for implementing some of the fitness functions. In \cref{tab:merge-methods} we outline the supported merging methods, while in \cref{tab:optimization_algorithms} we outline the currently available evolutionary algorithms.

The \mergenetic{} library is divided into distinct modules that reflect the core stages of evolutionary model merging: (i) defining the workflow (Python API, CLI, GUI), (ii) performing the merge (\code{Merger}), (iii) formulating the optimization problem (\code{Optimization}), (iv) evaluating merged models (\code{Evaluator}), and (v) orchestrating the evolution loop (\code{Searcher}). Below, we briefly describe each module and link it to the broader system design.

\subsection{Python API, CLI, and GUI}
\label{sec:api_cli_gui}

\paragraph{Python API.}
Figure \ref{fig:example_code} provides an example usage of the API.
The \code{Searcher} and \code{Problem} classes form the core of the Python API. Users can instantiate an optimization \emph{problem} (e.g., merging multiple language models), select an algorithm from \pymoo{}, and call \code{searcher.search()} to launch the evolutionary procedure. A typical workflow involves:
\begin{enumerate}
    \item Defining an \emph{evaluation dataset} and relevant \emph{performance metric}.
    \item Instantiating a \code{Merger} to specify how weights are combined.
    \item Passing these to a \code{MergingProblem} class, describing the evolutionary search space and objectives.
    \item Choosing a \code{GeneticAlgorithm} (e.g., \method{NSGA-II}, \method{GA}, \method{DE}) from \pymoo{}.
    \item Running the search, then optionally calling \code{.test()} on the best solutions.
\end{enumerate}

\paragraph{CLI.}
For users who prefer a command-line approach without manually writing scripts, the \mergenetic{} CLI is invoked via:
\begin{quote}
\small
\texttt{python mergenetic.py --eval-method <lm-eval|custom> --merge-type <single|multi>}
\end{quote}
Internally, it launches an interactive wizard to guide users through selecting \emph{models}, \emph{tasks}, \emph{algorithms}, and \emph{merging methods}. The CLI can handle four main modes: single- or multi-language merges, each with either \lmeval{} or \emph{custom} evaluations. By abstracting away many details, the CLI lets users prototype merges quickly with no code.

\begin{figure*}
    \centering
    \includegraphics[width=\textwidth]{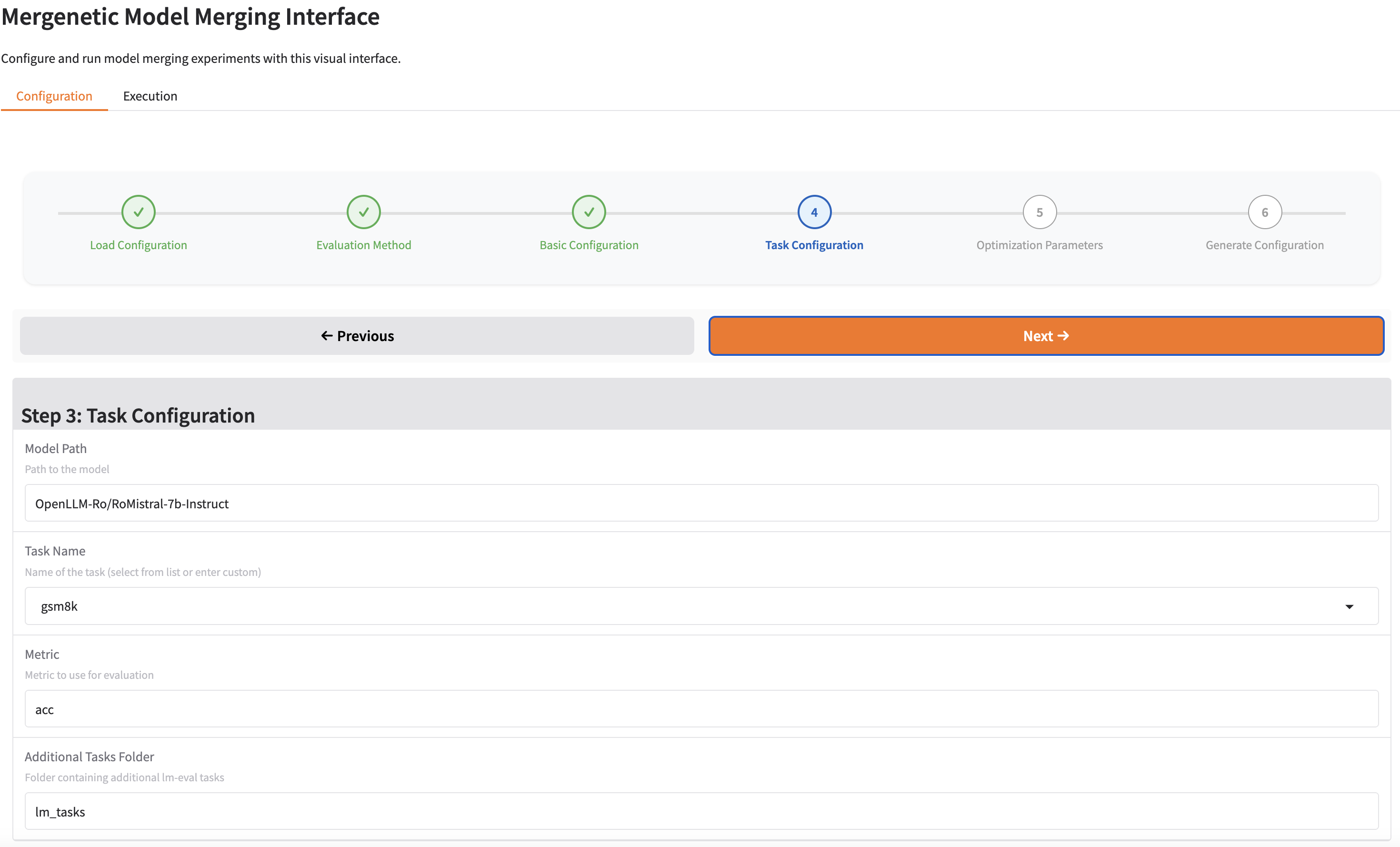}
    \caption{Screenshot of the Gradio-based GUI described in \cref{sec:api_cli_gui}. The user is guided through a step-by-step process to define every ingredient of the evolutionary merging pipeline.}
    \label{fig:GUI}
\end{figure*}

\paragraph{GUI.}
A Gradio\footnote{\href{https://github.com/gradio-app/gradio}{https://github.com/gradio-app/gradio}}-based \cite{abid2019gradio} graphical interface provides a further layer of accessibility, especially for non-technical users or demonstration purposes (See Fig. \ref{fig:GUI}). It reuses the same core configuration concepts but wraps them in a step-by-step wizard: (1) load base model(s), (2) specify tasks/languages, (3) set evolutionary parameters, and (4) run merging with real-time logs. The GUI allows merging without coding.

\subsection{Core components}
We describe here the core components.
\subsubsection{Merger}
\label{sec:merger}
The \code{Merger} module handles the core weight-combination logic by interfacing with \mergekit{}. Each merger class (e.g., \code{SlerpMerger}, \code{TiesDareMerger}, \code{TaskArithmeticMerger}) generates a YAML configuration specifying the base checkpoints, interpolation method, and merge coefficients. This configuration is passed to \mergekit{}, which performs the actual merging and produces a new model checkpoint. The merger supports both standard and multi-model merges, including advanced strategies like \method{TIES} combined with \method{DARE} \cite{ties,yu2024language}. Additionally, \mergenetic{} manages GPU memory during the evolutionary search, helping avoid out-of-memory errors. During optimization, the evolutionary algorithm proposes weight combinations, which the merger translates into actual models ready for evaluation.

\subsubsection{Optimization}
\label{sec:opt}

At the core of \mergenetic{}, the optimization module casts model merging as a \emph{black-box optimization} problem. The decision variables correspond to the targeted parameters from the merging configuration file (the genotype), such as the interpolation or pruning coefficients. Objective functions define the fitness criteria to be optimized, such as accuracy, perplexity, or other task-specific metrics.

The \code{MergingProblem} class define how to:
\begin{enumerate*}[label=(\roman*)]
    \item Convert a genotype to a merged model (by calling the \code{Merger}).
    \item Evaluate the merged model on a dataset, via an \code{Evaluator}.
    \item Return the resulting fitness or multi-objective scores to the algorithm.
\end{enumerate*}

Using \pymoo{}~\cite{pymoo}, \mergenetic{} supports a variety of \textbf{single-} or \textbf{multi-objective} methods. Single-objective approaches optimize \emph{one} metric (e.g., cross-lingual accuracy), while multi-objective strategies (e.g., NSGA-II) can simultaneously balance multiple metrics like \emph{math accuracy} vs.\ \emph{general fluency}.

\subsubsection{Evaluator}
\label{sec:evaluator}

Evaluators compute a merged model's performance on the chosen task(s). In \mergenetic{}, they appear both as \emph{direct evaluators} (e.g., running on a small dataset) or as \emph{IRT-based estimators} using anchors \cite{merge3}. We highlight two broad categories:

\paragraph{LM-Eval-Harness Evaluators.}
\mergenetic{} can natively call out to the \lmeval{}~\cite{lm-eval} library, passing the merged checkpoint and a chosen benchmark (e.g., ARC, GSM8K). This approach covers many standard tasks and yields consistent comparisons. However, it can be relatively expensive if one repeatedly evaluates large datasets on many candidate merges. To offset this problem, \mergenetic{} wrap \lmeval{} and allow explicit subsamples through the plug-and-play \code{ConfigPE}, which allows to subsample such without the need to instantiate a new \lmeval{} config file.

\paragraph{Custom Evaluators.}
Users can alternatively define their own logic for computing correctness---e.g., \code{MultilingualMathFGEvaluator} that checks whether the final extracted number is correct \emph{and} in the target language. Or a \code{MultipleChoiceEvaluator} that compares the chosen letter (A, B, C, D) to the ground truth. These evaluators easily allow advanced users to combine partial correctness checks with domain constraints (e.g., ``the predicted chemical formula must be balanced'').

\subsubsection{Searcher}
\label{sec:searcher}

Finally, the \code{Searcher} orchestrates the evolutionary loop: it begins with the \textbf{initialization} of a population of random genotypes (weight vectors), followed by \textbf{merging/evaluation}, where each genotype is merged into a checkpoint and scored on user-specified tasks/datasets. Then comes \textbf{selection/variation}, where parent genotypes are chosen based on fitness and modified via crossover and mutation to produce children. Steps 2 and 3 repeat for $T$ generations in the main \textbf{loop}.
Therefore, the \code{Searcher} class essentially wraps all these elements (\code{Problem}, \code{Merger}, \code{Evaluator}, \pymoo{} \code{algorithm}) in an easy-to-use API.

During the search process, intermediate results (population genotypes, partial solutions, logs) are stored in \code{CSV} or \code{JSON}, facilitating real-time monitoring. At completion, \texttt{test()} re-merges the best solutions and evaluates them on an unseen test set to quantify final performance.

\section{Case Studies}
To demonstrate the capabilities of the \mergenetic{} library, we reproduce here two evolutionary model merging pipelines: MERGE$^3$ \cite{merge3} and EvoLLM-JP \cite{sakana}.

\begin{figure}
    \centering
    \includegraphics[width=\linewidth]{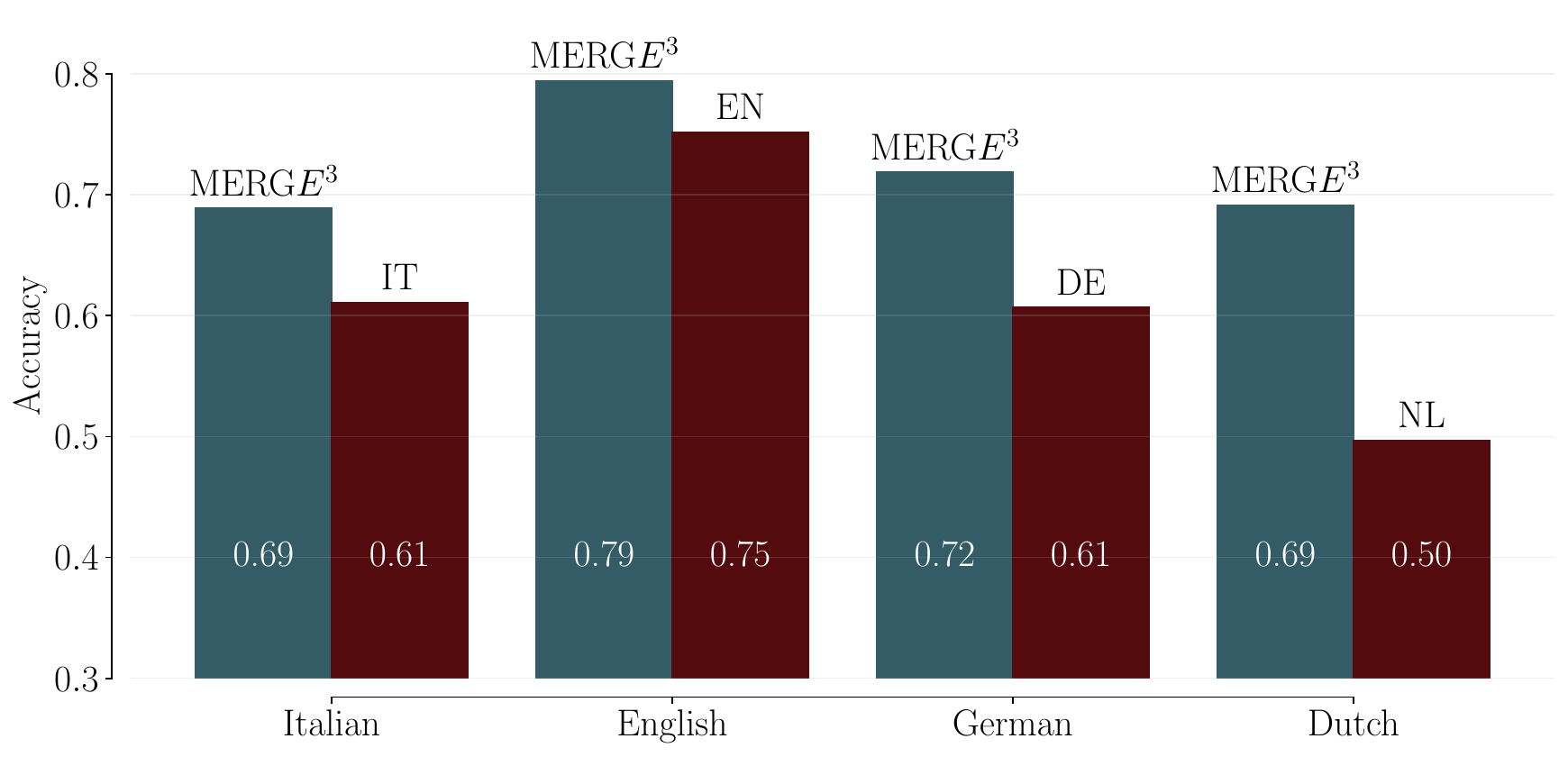}
    \caption{Evolving a multi-lingual model spanning Italian, English, German and Dutch.}
    \label{fig:multilingual-exp}
\end{figure}

\subsection{Evolving a multi-lingual model}
We demonstrate how \mergenetic{} can be used to merge individually fine-tuned models for four languages --- Italian, English, German, and Dutch --- into a single multilingual model. This setup formulates the objective function as explicitly multi-task, assigning one evaluation metric per language to promote balanced cross-lingual performance. Details on the specific models used per language are provided in \Cref{app:multilingual}. As shown in \cref{fig:multilingual-exp}, the merged model consistently outperforms each of its language-specific constituents, achieving up to a 19\% accuracy gain on the \texttt{ARC-Challenge} benchmark~\cite{ARC}. Notably, it surpasses all endpoints across the board, highlighting the effectiveness of evolutionary merging in facilitating positive knowledge transfer across languages.

\subsection{Cross-lingual transfer}
\begin{figure}
    \centering
    \includegraphics[width=\linewidth]{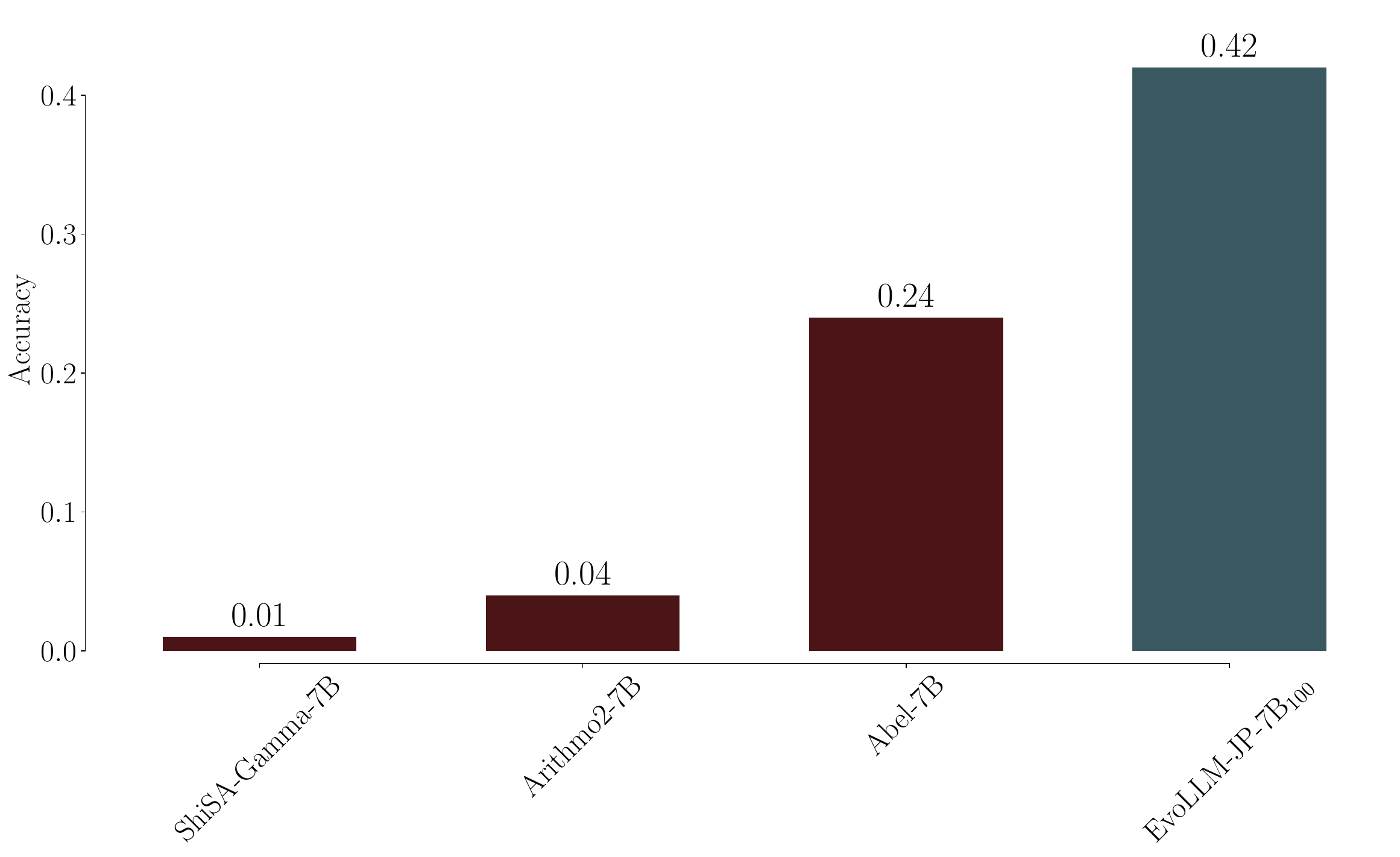}
    \caption{Cross-lingual transfer of math solving capabilities from English to Japanese.}
    \label{fig:crosslingual-exp}
\end{figure}

To showcase the ability of \mergenetic{} to support cross-lingual skill transfer, we merge a math-specialized English model with a Japanese fine-tuned version of \texttt{Mistral-7B} \cite{mistral}, and evaluate the result on the Japanese translation of the \texttt{GSM8K} dataset \cite{gsm8k}. This experiment follows the general setup proposed by \citet{sakana}, using a subset of $100$ samples for the fitness evaluation instead of the full dataset. As shown in \cref{fig:crosslingual-exp}, the merged model achieves a 10-20\% accuracy improvement over each of its individual components, demonstrating effective cross-lingual transfer enabled by evolutionary merging.

\section{Conclusions}

\mergenetic{} bridges the gap between cutting-edge evolutionary model merging and practical usability on consumer hardware. By combining flexible merging strategies, diverse evolutionary algorithms, and lightweight fitness approximators, it empowers researchers and practitioners to explore high-quality model compositions without requiring large-scale infrastructure. Through its Python API, CLI, and GUI, \mergenetic{} supports both systematic experimentation and user-friendly workflows. We hope the library will serve as a stepping stone for future research in multilingual, multi-task, and efficient evolutionary model merging, and invite the community to build upon and extend its capabilities.

\clearpage

\section*{Limitations}

While \mergenetic{} significantly lowers the entry barrier for evolutionary model merging, several limitations remain:

\paragraph{Dependence on Existing Fine-Tuned Models.}
Model merging requires access to pre-trained or fine-tuned base models with relevant capabilities (e.g., math reasoning, language-specific fluency). As such, the technique currently cannot be directly applied to extremely low-resource languages or domains where such models are unavailable. This limits its immediate applicability in truly zero-resource settings. Future work could explore integrating lightweight fine-tuning or retrieval-based augmentation prior to merging to alleviate this dependency.

\paragraph{Hardware Requirements.}
Although \mergenetic{} is designed for consumer-grade GPUs, it still requires relatively high-tier hardware (e.g., NVIDIA RTX 2080 or better) due to the size of language models involved and the need to load and evaluate them during evolution. Most laptops or low-memory GPUs may not have sufficient VRAM to support repeated merging and evaluation steps. We see this as a broader limitation of current LLM infrastructure and hope that advances in model quantization, sparse evaluation, and efficient loading techniques will further democratize access to frontier AI tools like \mergenetic{}.

\bibliography{custom}

\clearpage

\appendix

\section{Additional Details}
\label{sec:appendix}

\subsection{Cross-Lingual Case Study Details}\label{app:crosslingual}

For the cross-lingual case study, we conduct evolutionary search on the Japanese subset of the MGSM dataset~\cite{mgsm}, a multilingual extension of GSM8K~\cite{gsm8k}. The final merged model is evaluated on the MGSM test set, following the evaluation protocol of \citet{sakana}. Unlike their setup, which used 1,069 search datapoints (the remaining part of GSM8k test set that was not included in MGSM), we use only a subset of 100 examples for computational efficiency. Our approach employs a single-objective evolutionary algorithm based on a Genetic Algorithm~\cite{dasgupta1997evolutionary}, incorporating a Simulated Binary Crossover (SBX) operator~\cite{sbx} for recombination and a Polynomial Mutation operator~\cite{sbx} for exploration. We set the population size to 25 and run the algorithm for 7 generations. Fitness and evaluation metrics are computed by extracting the final numeric answer using a regular expression and verifying both the mathematical correctness and the linguistic accuracy of each response. Language identification is performed using the method described in~\cite{ft1}. Only responses that are both mathematically and linguistically correct are considered valid. The models evaluated in this experiment include \texttt{Arithmo2-Mistral-7B}, \texttt{Abel-7B-002}, and \texttt{shisa-gamma-7b-v1}.

\subsection{Multilingual case study details}\label{app:multilingual}
For the multilingual case study, we perform evolutionary model merging across four languages ---Italian, Dutch, German, and English --- using the translated \texttt{ARC} dataset from the Hugging Face repository \cite{thellmann2024crosslingual}\footnote{\url{https://huggingface.co/openGPT-X/arcx}}. We employ a multi-objective optimization setup with NSGA-II~\cite{nsga-ii}, configuring the evolutionary process with a population size of 25 and 7 iterations. As the merging strategy, we use a combination of \method{TIES} and \method{DARE}. The fitness and test evaluations are performed by extracting the final answer choice (A, B, C, or D) from the model's output using a regular expression. For each language, we use a reduced dataset of 20 translated examples from \texttt{ARC} to compute fitness scores, keeping the process efficient and GPU-friendly. The models used in this experiment include \texttt{Mistral-Ita-7B}, \texttt{GEITje-7B-ultra}, \texttt{leo-mistral-hessianai-7B}, and the base model \texttt{Mistral-7B-v0.1}.

\begin{table*}[t]
\footnotesize
\centering
\resizebox{\textwidth}{!}{%
\begin{tabular}{l|lccl}
\toprule
\textbf{Algorithm} & \textbf{Class} & \textbf{Obj.} & \textbf{Constr.} & \textbf{Description} \\
\midrule
Genetic Algorithm & GA & single & \checkmark & Customizable evol. operators for broad problem categories \\

Differential Evol. & DE & single & \checkmark & Variants for continuous global optimization \\

BRKGA & BRKGA & single & \checkmark & Advanced variable encoding for combinatorial opt. \\

Nelder Mead & NelderMead & single & \checkmark & Point-based algorithm using simplex operations \\

Pattern Search & PatternSearch & single & \checkmark & Iterative approach with exploration patterns \\

CMAES & CMAES & single & & Model-based sampling from dynamic normal distribution \\

Evol. Strategy & ES & single & & Real-valued optimization strategy \\

SRES & SRES & single & \checkmark & ES with stochastic ranking constraint handling \\

ISRES & ISRES & single & \checkmark & Improved SRES for dependent variables \\

NSGA-II & NSGA2 & multi & \checkmark & Non-dominated sorting and crowding \\

R-NSGA-II & RNSGA2 & multi & \checkmark & NSGA-II with reference points \\

NSGA-III & NSGA3 & many & \checkmark & NSGA-II for many-objective problems \\

U-NSGA-III & UNSGA3 & many & \checkmark & NSGA-III optimized for fewer objectives \\

R-NSGA-III & RNSGA3 & many & \checkmark & NSGA-III with aspiration points \\

MOEAD & MOEAD & many & & Multi-objective optimization via decomposition \\

AGE-MOEA & AGEMOEA & many & & Estimates Pareto-front shape instead of crowding \\

C-TAEA & CTAEA & many & \checkmark & Sophisticated constraint-handling for many objectives \\

SMS-EMOA & CTAEA & many & \checkmark & Uses hypervolume during environmental survival \\

RVEA & RVEA & many & \checkmark & Reference direction with angle-penalized metric \\
\bottomrule
\end{tabular}
}
\caption{Supported Optimization Algorithms and their description.}
\label{tab:optimization_algorithms}
\end{table*}

\subsection{Supported evolutionary algorithms}
\Cref{tab:optimization_algorithms} lists all the evolutionary algorithms provided by \pymoo{} and hence supported in \mergenetic{}, stating whether they are single- or multi-objective and if they allow constraints to be defined, along with a brief description.

\subsection{Performance Estimator}

To reduce the computational cost associated with evaluating the fitness of candidate models during evolutionary merging, the \mergenetic{} library supports estimator-based approximations inspired by \citet{merge3} and \citet{tinybenchmarks}. These methods allow us to estimate model performance using a reduced subset of the evaluation dataset, significantly accelerating the evolution process without sacrificing accuracy.

In particular, \mergenetic{} provides implementations of both standard and merging-specific IRT-based estimators, which leverage latent ability inference to approximate full-dataset correctness. These estimators vary in their assumptions and complexity, offering a trade-off between computational efficiency and estimation fidelity.

Table~\ref{tab:estimators} provides an overview of the currently supported estimators, including a brief description and a qualitative rating of their performance.

\begin{table*}
\centering
\begin{tabular}{l p{8cm} c}
\toprule
\textbf{Estimator} & \textbf{Description} & \textbf{Performance} \\
\midrule
Random & Baseline estimator using random sample correctness. Simple but noisy and unreliable. & $\star\star$ \\
P-IRT & Standard Item Response Theory estimator, uses subset to estimate ability, not tailored for merging. & $\star\star\star$ \\
GP-IRT & Generalized P-IRT with better smoothing but still not designed for merging. & $\star\star\star$ \\
MP-IRT & MERGE3's merged-performance IRT estimator assuming linear combination of abilities. & $\star\star\star \star$ \\
GMP-IRT & Generalized version of MP-IRT, combines predictions and observations with learned weights. & $\star\star\star\star$ \\
Full Dataset & Ground truth performance by running evaluation on the full dataset. & $\star\star\star\star\star$ \\
\bottomrule
\end{tabular}
\caption{Comparison of different performance estimators.}
\label{tab:estimators}
\end{table*}

\subsection{License}
The library is licensed with the \texttt{Apache 2.0} license. This means that it can be freely used, modified, and redistributed by anyone, including for commercial purposes. The license is designed to promote widespread adoption by offering a permissive legal framework that imposes minimal restrictions on end users. Developers are allowed to modify the source code and distribute derivative works under different terms, provided that the original license and copyright notice are retained.
Mergenetic builds upon two key dependencies: \pymoo{} and \mergekit{}. The former is distributed under the same \texttt{Apache 2.0} license as \mergenetic{}, ensuring compatibility and permissive use. However, \mergekit{} introduces additional licensing constraints in versions beyond \texttt{v0.1.0}. Specifically, it adopts a Business Source License, which restricts production use based on organizational scale and revenue. Users intending to deploy \mergenetic{} for commercial purposes are advised to review these terms carefully and install a version of \mergekit{} that aligns with their intended usage scenario. If unrestricted commercial use is required, it is recommended to use version \texttt{0.1.0} of \mergekit{}, which remains under the Apache 2.0 license, or to contact the licensor for alternative licensing arrangements.

\end{document}